\documentclass{article}
\usepackage{spconf}
\usepackage{times}
\usepackage{amsmath}
\usepackage{graphicx}
\usepackage{hyperref}
\usepackage{epsfig}
\usepackage{amssymb}
\usepackage{cite}
\usepackage{subfigure}
\usepackage{xspace}
\usepackage{xcolor}
\usepackage{varwidth}

\def\eg{\emph{e.g.}\xspace} 
\def\ie{\emph{i.e.}\xspace}

\DeclareMathOperator*{\argmin}{arg\,min}

\newcommand{\norm}[1]{{\left\Vert #1 \right\Vert}}

\title{Convolutional Neural Network Pruning to Accelerate Membrane Segmentation in Electron Microscopy}
\name{Joris Roels\textsuperscript{1,3}, Jonas De Vylder\textsuperscript{1}, Jan Aelterman\textsuperscript{1}, Yvan Saeys\textsuperscript{2,3}, Wilfried Philips\textsuperscript{1}
\thanks{
This research has been made possible by the Agency for Flanders Innovation \& Entrepreneurship (VLAIO) and the IMEC BAHAMAS project (\url{http://www.iminds.be/en/projects/2015/03/11/bahamas}). We would like to thank Saskia Lippens\textsuperscript{3} and Michiel Krols\textsuperscript{3} for the provided electron microscopy data and annotations. }}
\address{
\textsuperscript{1}Ghent University - TELIN - IMEC, \\
Sint-Pietersnieuwstraat 41, B-9000 Ghent, Belgium \\
\textsuperscript{2}Ghent University - Department of Internal Medicine, \\
De Pintelaan 185, B-9000 Ghent, Belgium \\
\textsuperscript{3}VIB - Center for Inflammation Research, \\
Technologiepark 927, B-9052 Ghent (Zwijnaarde), Belgium \\
}
\begin{document}
%
\maketitle
\begin{abstract}
Biological membranes are one of the most basic structures and regions of interest in cell biology. In the study of membranes, segment extraction is a well-known and difficult problem because of impeding noise, directional and thickness variability, etc. Recent advances in electron microscopy membrane segmentation are able to cope with such difficulties by training convolutional neural networks. However, because of the massive amount of features that have to be extracted while propagating forward, the practical usability diminishes, even with state-of-the-art GPU's. A significant part of these network features typically contains redundancy through correlation and sparsity. In this work, we propose a pruning method for convolutional neural networks that ensures the training loss increase is minimized. We show that the pruned networks, after retraining, are more efficient in terms of time and memory, without significantly affecting the network accuracy. This way, we manage to obtain real-time membrane segmentation performance, for our specific electron microscopy setup. 
\end{abstract}
\section{INTRODUCTION}
\label{sec:intro}
State-of-the-art imaging devices such as electron microscopy (EM) allow life science researchers to study biological structures at high resolution. Due to advanced EM imaging, researchers have found e.g. a correlation between mitochondria and endoplasmic reticula (ER) on the one hand and diseases such as cancer and Parkinson on the other hand \cite{Zong2016,Tsujii2015}. Consequently, high throughput image analysis of these structures has gained a lot of attention the last years. 

Because of the big data volumes and complex image content, this analysis requires computationally efficient and high-quality segmentation algorithms. However, state-of-the-art segmentation algorithms typically require more computational effort in order to capture more feature variability or modeling. Advanced and expensive 3D EM devices are therefore not fully exploited as long as real-time and high-quality segmentation (jointly) is not possible.

\begin{figure}
	\label{fig:em}
	\centering
	\includegraphics[width=0.95\linewidth]{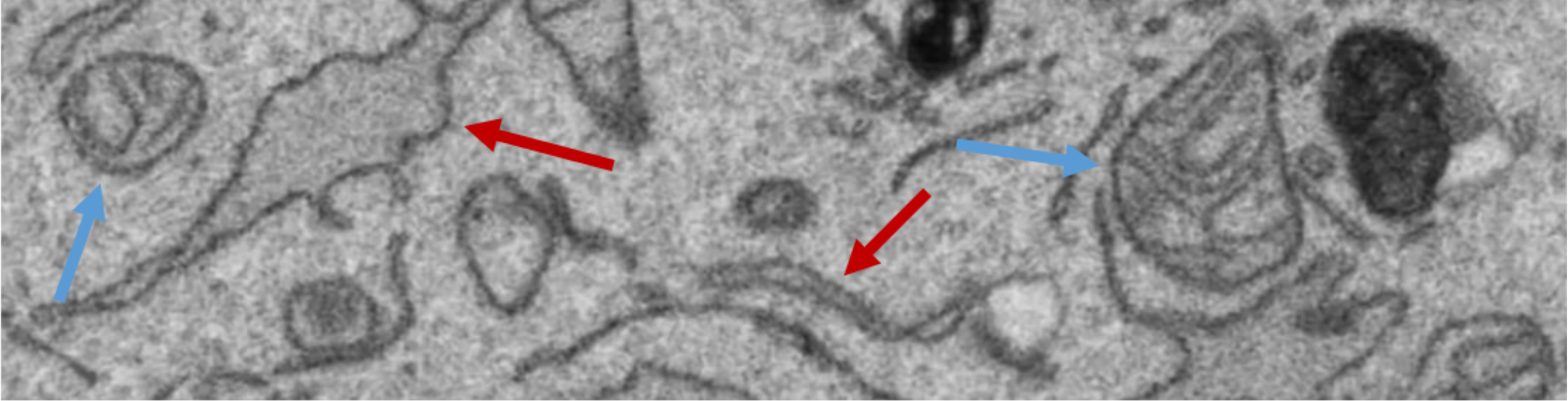}
	\caption{Electron micrograph showing several mitochondria and endoplasmic reticula (ER), indicated in blue and red arrows, respectively. }
\end{figure}

Generally speaking, mitochondria and ER are delineated by membranes with the same visual characteristics (Fig. \ref{fig:em}). We propose a membrane extraction method based on convolutional neural networks that can help segmenting complete mitochondria and ER in Sec. \ref{sec:segmentation}. In order to improve the practical usablity, we prune and retrain the original network such that the most relevant features remain present and redundancy is reduced (Sec. \ref{sec:feature-selection}). The obtained networks are more efficient in terms of runtime and memory requirements, without significantly affecting the original accuracy (Sec. \ref{sec:results}). 

\begin{figure*}[t]
	\begin{center}
		\centerline{\includegraphics[width=\textwidth]{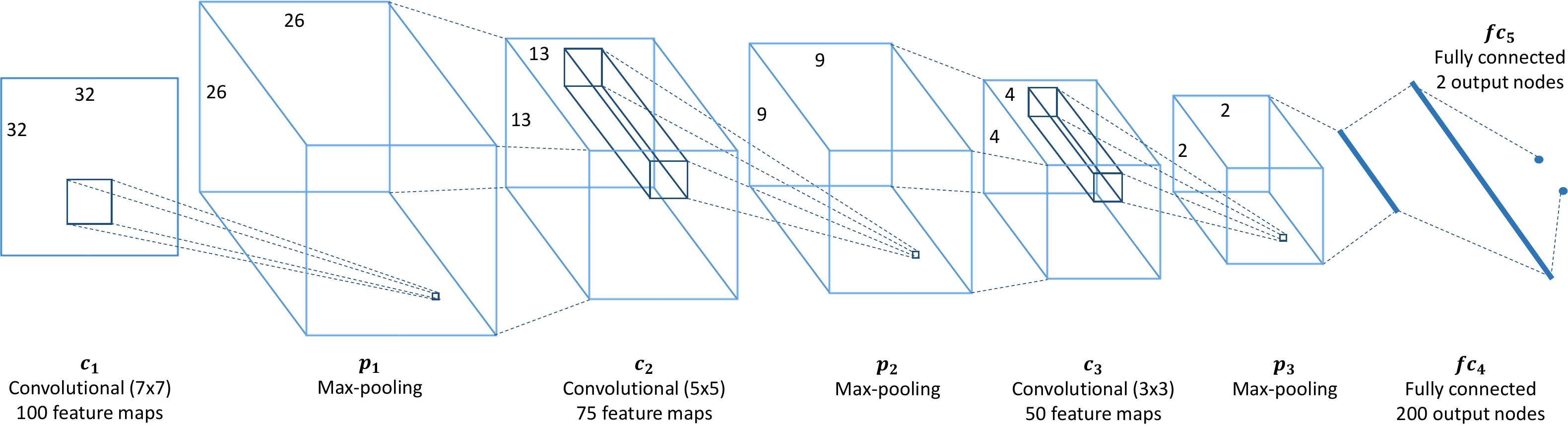}}
		\caption{Proposed network architecture}
		\label{fig:network}
	\end{center}
\end{figure*} 

\section{MEMBRANE SEGMENTATION IN EM}
\label{sec:segmentation}

\subsection{RELATED WORK}
\label{sec:related-work}
Mitochondria and ER segmentation has been a challenging research topic in recent literature \cite{Ghita2014,Ohta2014}. Since their corresponding outer membrane has the same visual characteristics (Fig. \ref{fig:em}), there have been various membrane segmentation proposals in literature \cite{Jurrus2009,Vazquez-Reina2009,Vu2008,Yang2009,Andres2008,Kaynig2010,Ciresan2012a,Masci2013,Wu2015a}. We differentiate between model-driven and machine learning based methods. 

Model-driven segmentation methods model the image and object of interest through energy functions by using a priori knowledge: \eg active contours \cite{Jurrus2009,Vazquez-Reina2009} and graph-based methods \cite{Vu2008,Yang2009}. Even though it is possible to obtain high quality results using these techniques, they tend to be highly parameter dependent and are hard to generalize because of the specific image and object assumptions that were made. 

The availability of annotated EM data \cite{Cardona2010} has allowed supervised machine learning to improve the field of EM membrane segmentation (\eg using random forests \cite{Andres2008,Kaynig2010}). Most of these techniques rely on greedy feature selection from the image, \eg intensity, edge, texture or shape-based. Recent convolutional neural network (CNN) approaches \cite{Ciresan2012a,Masci2013,Wu2015a} have allowed researchers to train relevant features directly from the labeled data and improve the state-of-the-art. 

CNN's extract a large amount of features by combining various layers of convolutions, making them computationally intensive. Even with the most advanced GPU's, this can become an issue. Therefore, pruning methods have been proposed \cite{Anwar2015,Han2015,Han2016}. These techniques remove feature maps, based on particle filters \cite{Anwar2015} or weight sparsity \cite{Han2015,Han2016}, and retrain the pruned network. In this work, we propose a pruning method based on training loss minimization, while maintaining the original network accuracy. Our method shows competitive performance, compared to state-of-the-art sparsity-based pruning. 

\subsection{PROPOSED NETWORK}
\label{sec:proposed-network}

Similar to most approaches \cite{Ciresan2012a,Masci2013,Wu2015a}, our proposed membrane segmentation algorithm follows a pixel-based classification strategy. For a given image $I:\mathbb{R}^2 \rightarrow \mathbb{R}^{+}$ (where $\mathbb{R}^{+}$ is the set of positive real numbers) we extract a local $n \times n$ patch $P_{i,j}$ centered around each pixel position $(i,j)$ (we observed an optimal trade-off between local information and computational effort for $n=32$). Each patch $P_{i,j}$ is propagated forward in an $8$-layer CNN in order to classify the center pixel $I(i,j)$ of the patch as being either a membrane pixel or not. 

Motivated by the architecture proposed in \cite{Ciresan2012a}, our network consists of three convolutional layers, denoted by $c_i$ ($i=1,2,3$), with stride $1$ and kernel size $7$, $5$ and $3$ (and respectively $100$, $75$ and $50$ output maps). Each convolutional layer $c_i$ is followed by a max-pooling layer $p_i$ with stride $2$ and kernel size $3$. The last two layers of the network are fully connected, consist of respectively $200$ and $2$ output maps and are denoted by $fc_i$ ($i=4,5$). A detailed overview is shown in Figure \ref{fig:network}. 

In order to train this network, we annotated mitochondria in a $3069 \times 2301 \times 689$ scanning electron microscopy (SEM) dataset (Fig. \ref{fig:em} shows a crop) and extracted the membrane pixels (approximately $300,000$ in total). We randomly annotated the same number of non-membrane pixels in the same dataset. This resulted in a training dataset $\{ P_i^{\text{train}} \}_{1 \leq i \leq N}$ of $N=300,000$ equally represented positive and negative membrane patches and a validation dataset $\{ P_i^{\text{val}} \}_{1 \leq i \leq N}$ of the same size. 

Given this setup, we minimized the loss function: 
\begin{equation}\label{eq:loss-function}
	\mathcal{L}(W) = \sum\limits_{i=1}^{N} f_W(P_i^{\text{train}}) + \lambda \norm{W}_2^2
\end{equation}
w.r.t. the network weights $W$, \ie the convolution, inproduct and bias variables. In Eq. \ref{eq:loss-function}, $f_W(P)$ denotes the entropy classification loss on patch $P$ w.r.t. the weights $W$. Note we also applied $L_2$ regularization in the optimization for the weights with its corresponding regularization parameter $\lambda \geq 0$. An estimation $\hat{W}$ for $W$ was obtained by stochastic gradient descent (SGD), using the Caffe framework \cite{Jia2014}. We obtained good convergence with a linearly decreasing learning rate initialized at $0.001$, a momentum of $0.9$ and regularization value of $\lambda=0.01$. 

\section{NETWORK FEATURE SELECTION}
\label{sec:feature-selection}

The estimated feature parameters $\hat{W}$ have many commonalities within each layer, resulting into redundant features that contribute little to the network performance. We propose a pruning method where feature parameters are discarded such that the training loss increase is minimized. 

Each convolutional and fully connected layer has a specific number of outputs that correspond to a specific feature (\ie a convolution kernel or an inproduct parameter). Consider the first layer ($c_1$), where that number of trained features is $n_1 = 100$. We compute the training loss increase in case feature $f \in [1,n_1]$ would be discarded from the network (\ie its corresponding parameters would be set to $0$). In a greedy sense, the feature that corresponds to the smallest training loss should be a good candidate to discard from the network: 
\begin{equation}\label{eq:loss-increase}
f_1 = \argmin_{f \in [1,n_1]} \mathcal{L}\left(\hat{W}_{\{f\}}\right),
\end{equation}
where $\hat{W}_F$ is equal to $\hat{W}$, except for the convolution (or inproduct, in case of full connections) and bias parameters that correspond to each feature $f \in F$ in the current layer. In our approach the loss was computed as the average loss on a randomly selected set of mini-batches. 

Similarly, assuming a set of features $\{f_1,\dots,f_l\}$ have already been removed from the first layer, we will select the next feature $f_{l+1}$ from the remaining ones in that layer as the one that results in a minimal training loss, when discarded: 
\begin{equation}\label{eq:loss-increase-further}
f_{l+1} = \argmin_{f \in [1,n_1] \backslash \{f_1,\dots,f_l\}} \mathcal{L}\left(\hat{W}_{\{f_1,\dots,f_{l},f\}}\right). 
\end{equation} 
This way, by recursion, we can determine a feature ordering $F_{1} = \left(f_1,\dots,f_{n_1}\right)$ such that features $f_i$ ($i=1,\dots,d$) are the best candidates (w.r.t. the training loss) to discard from the network if we want to prune $d$ output maps in the first layer. Note that, assuming a specific subset of features has been removed from layers $1$ through $m$, it is possible to obtain a similar ordering $F_{m+1}$ for layer $m+1$. This way, we obtain a feature ordering for all layers except for the last layer $fc_5$, since the network output should obviously remain $2$ class probabilities. Note it is also possible to derive this feature ordering over the complete network, instead of layer per layer. The reason why we did not chose to do so, is to reduce the computational workload during pruning and to see how different layers respond to pruning. We obtained the best results by first pruning the bottom layers, and proceeding as we go deeper in the network. We think this is due to the fact that bottom layer adjustments might affect the top layers more than the other way around. 

After the pruning procedure, we retrained the network in order to optimize the smaller architecture. The previous obtained parameters were used as an initialization, which results in less required retraining iterations. We also had to decrease the learning rate since the network parameters are already in a near-optimal state.

\section{RESULTS \& DISCUSSION}
\label{sec:results}
We implemented our proposed neural network (denoted by N) and $7$ pruned variants of the network (denoted by N$_i$). In the first network N$_1$ we pruned $10$, $0$, $10$ and $50$ feature maps from the $c_1$, $c_2$, $c_3$ and $fc_4$ layers, respectively, of the initial network. These numbers were chose such that the network accuracy was not negatively influenced. Networks N$_2$ through N$_5$ differ from N$_1$ in exactly one layer, so as to see how different layers respond to pruning. Network N$_6$ combines the latter networks by selecting the minimal amount of feature maps in each layer. Lastly, we pruned an extremely large amount of feature maps in the N$_7$ network (more than $90\%$ of the parameters were dropped) in order to see how the network performance is influenced. A detailed description of the networks is given in Table 1. We implemented, trained and validated our CNN's using an NVIDIA GeForce GTX 1080 GPU and an Intel Core i7-3930K CPU @ 3.20GHz. 

\begin{figure}[t]
	\centering
	\begin{minipage}{0.48\linewidth}
		\includegraphics[width=\textwidth]{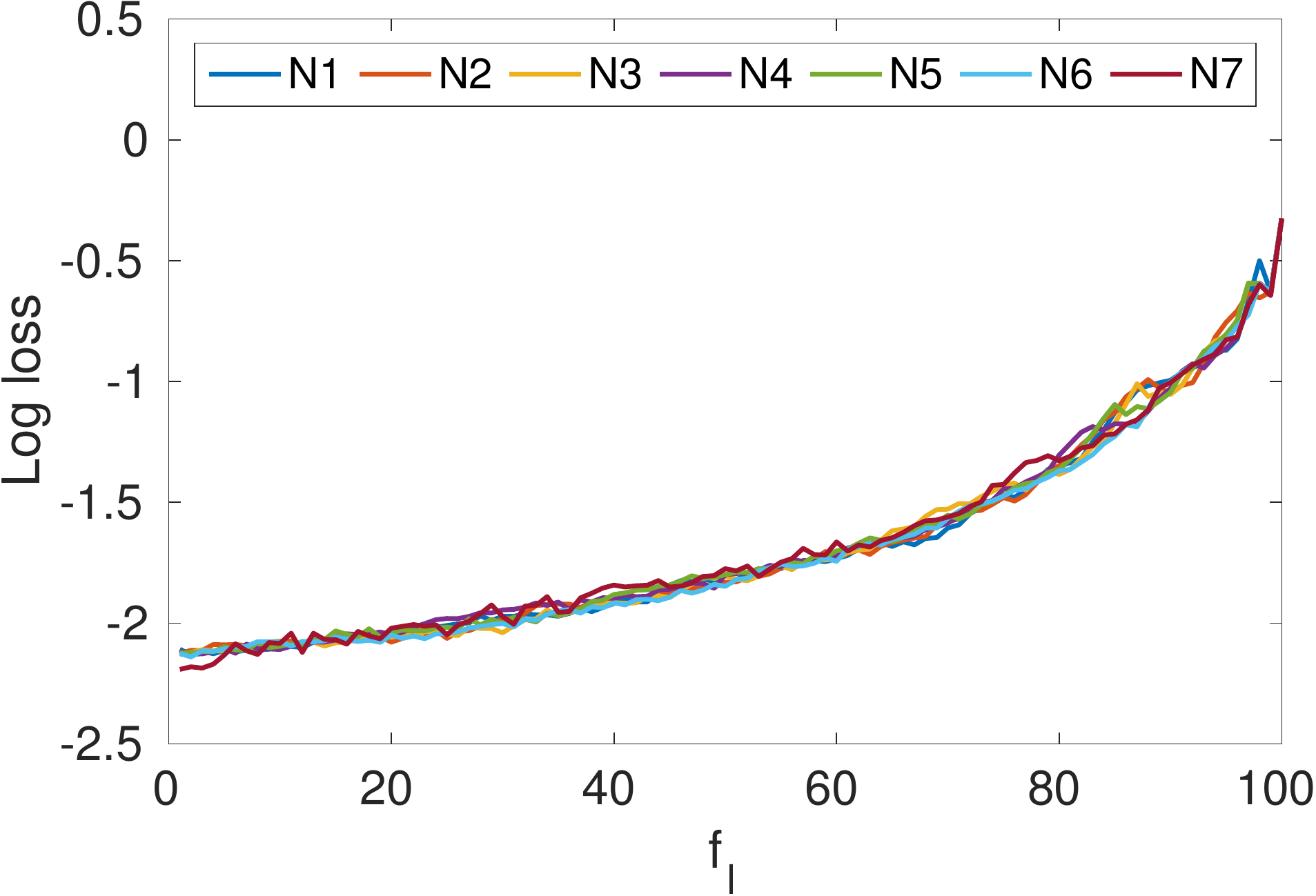}
		\centerline{(a) $c_1$ layer}
	\end{minipage}
	\begin{minipage}{0.48\linewidth}
		\includegraphics[width=\textwidth]{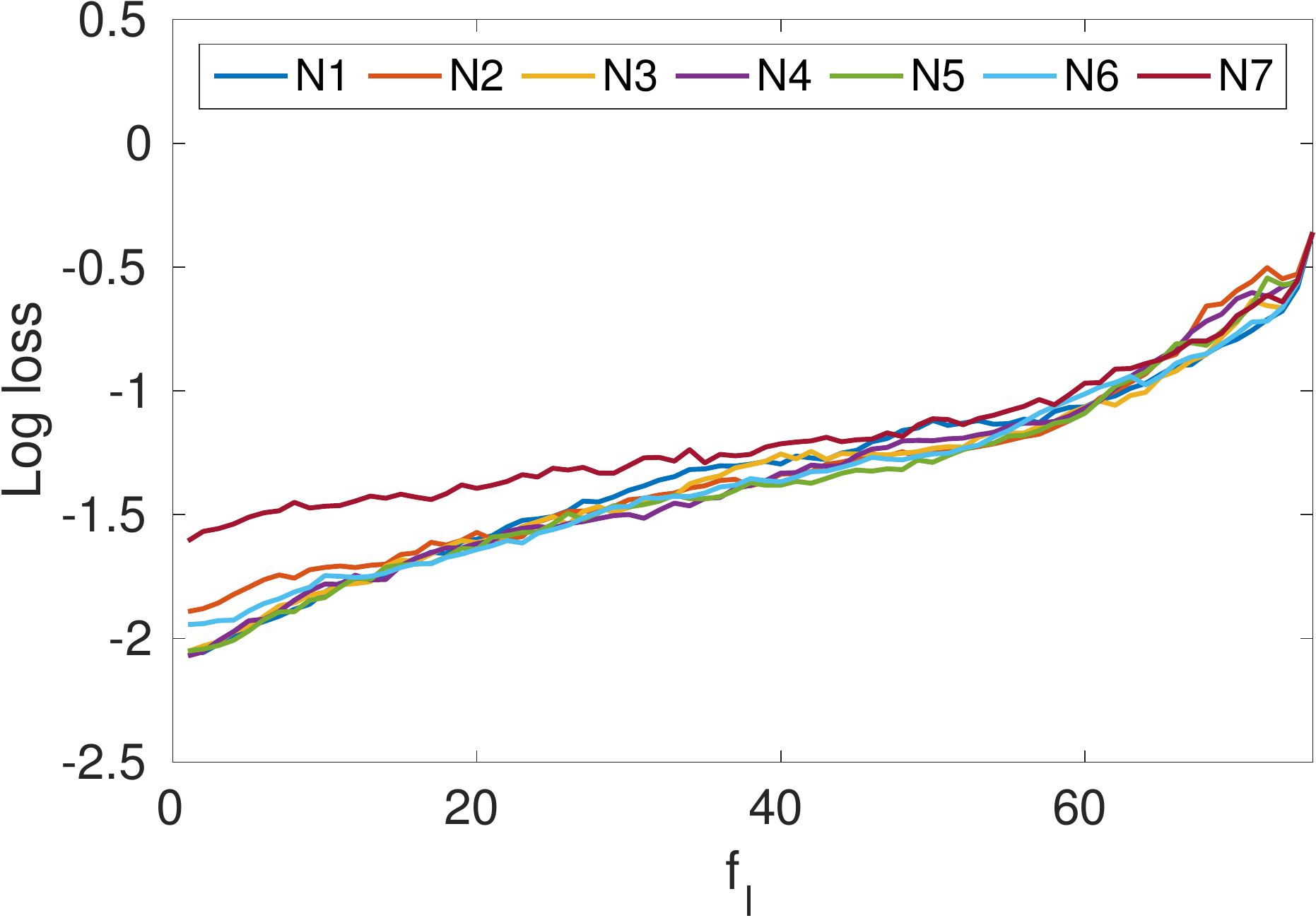}
		\centerline{(b) $c_2$ layer} 
	\end{minipage}
	\begin{minipage}{0.48\linewidth}
		\includegraphics[width=\textwidth]{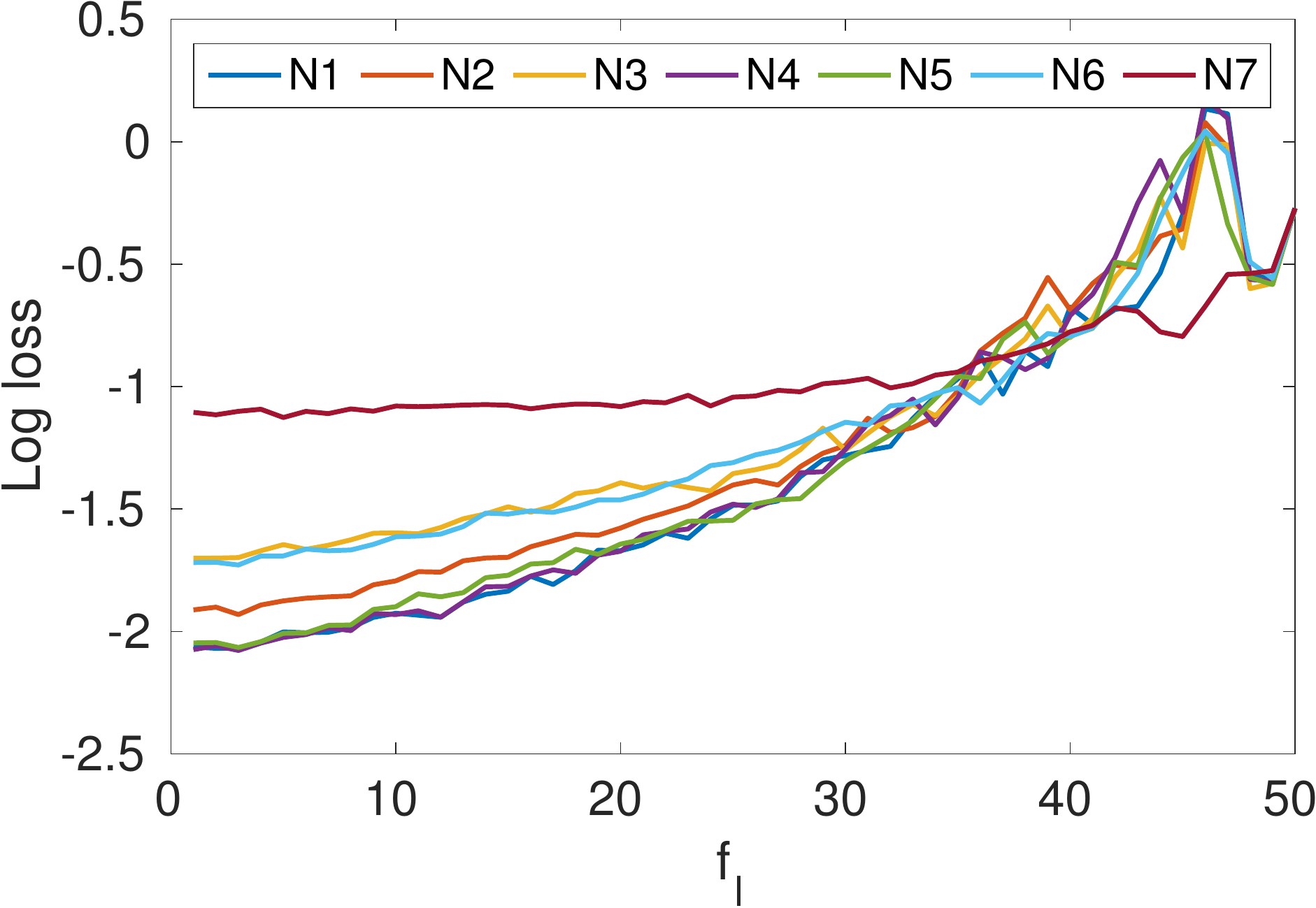}
		\centerline{(c) $c_3$ layer}
	\end{minipage}
	\begin{minipage}{0.48\linewidth}
		\includegraphics[width=\textwidth]{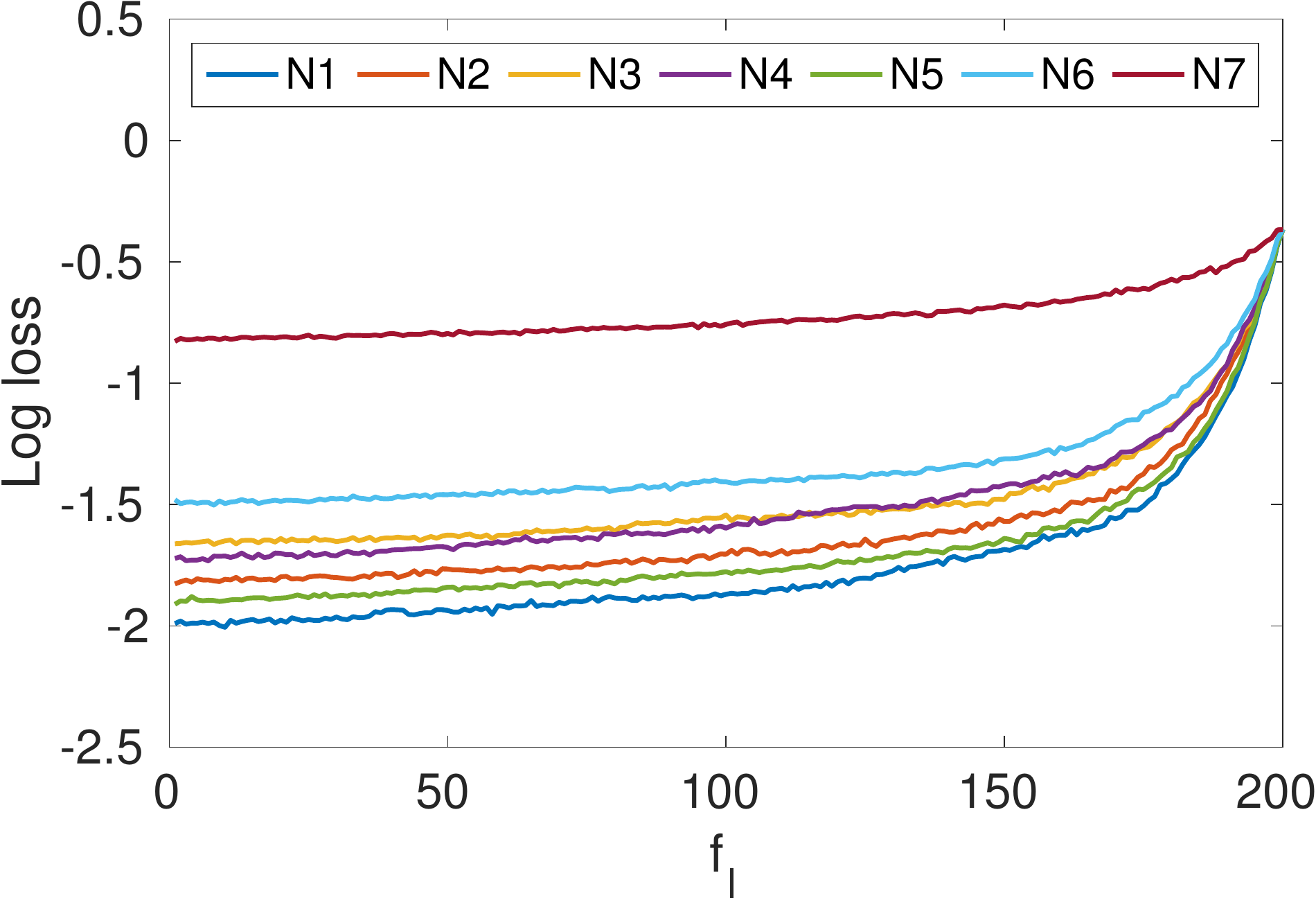}
		\centerline{(d) $fc_4$ layer}
	\end{minipage}
	\caption{Plots of the ordered features $F_{m}$ against their corresponding loss defined by Eq. \ref{eq:loss-increase-further} for each layer. }
	\label{fig:loss-increase}
\end{figure}

Fig. \ref{fig:loss-increase} plots the ordered features $f_l$ obtained from the pruning procedure against the corresponding loss (the minimum value in Eq. \ref{eq:loss-increase} and \ref{eq:loss-increase-further}) if $f_1, \dots, f_l$ would be discarded from the different networks for each layer. As to be expected, the loss increase for the first layer is the same for all networks (the small deviations are due to random batch selection in SGD). In layer $c_2$, we observe a loss increase for the N$_2$, N$_6$ and N$_7$ network since output maps have been removed from the $c_1$ layer. However, note that this loss difference becomes similar to the other layers if we were to discard more than $30$ features from layer $c_2$. A similar remark yields for networks N$_2$, N$_3$, N$_6$ and N$_7$ in layer $c_3$. Note also that the steepness of the curves is different across all layers, indicating some layers might need more feature maps or vice versa. 

\begin{table}
	\caption{Number of output feature maps of the original proposed network N and the number of output maps that were kept in the pruned networks N$_i$. }
	\centering 
	\begin{tabular}{|l|c|c|c|c|c|c|c|c|}
		\hline 
		 		& N 	& N$_1$	& N$_2$ & N$_3$ & N$_4$ & N$_5$ & N$_6$ & N$_7$ \\ \hline
		 $c_1$ 	& 100 	& 90 	& 65 	& 90 	& 90 	& 90 	& 65 	& 30 	\\ \hline
		 $c_2$ 	& 75 	& 75 	& 75 	& 60 	& 75 	& 75 	& 60 	& 20 	\\ \hline
		 $c_3$ 	& 50 	& 40 	& 40 	& 40 	& 30 	& 40 	& 30 	& 10 	\\ \hline
		 $fc_4$ & 200 	& 150 	& 150 	& 150 	& 150 	& 110 	& 110 	& 10 	\\ \hline
	\end{tabular}
	\label{tab:network-results2}
\end{table}

\begin{table}
	\caption{Performance comparison of the original network N and the pruned networks N$_i$. We specify the validation accuracy $A$ (for both our proposed loss-based pruning method and the one proposed in \cite{Han2015}), required segmentation time $T$ (in seconds) for one $3069 \times 2301$ slice of our dataset, percentage of pruned parameters $\Delta P$ and network memory usage $M$ for a batch size of $1$ (in kilobytes). }
	\centering 
	\begin{tabular}{|l|c|c|c|c|c|}
		\hline 
			  & $A$ (prop.) 	& $A$ (\cite{Han2015}) 		& $T$ (s) 	& $\Delta P$ 	& $M$ (kB) \\ \hline
		N 	  & 93.18 			& 93.18						& 114.7 	& 0.0 			& 654 \\ \hline
		N$_1$ & \textbf{93.34} 	& 93.08 					& 104.5 	& 12.7 			& 594 \\ \hline
		N$_2$ & \textbf{93.33} 	& 93.07 					& 83.5 		& 33.0 			& 445 \\ \hline
		N$_3$ & \textbf{93.29} 	& 93.10 					& 86.3 		& 29.3 			& 583 \\ \hline
		N$_4$ & 92.91 			& \textbf{93.26}			& 104.5 	& 16.2 			& 593 \\ \hline
		N$_5$ & \textbf{93.29} 	& 93.24 					& 104.4 	& 13.5 			& 593 \\ \hline
		N$_6$ & 92.87 			& \textbf{93.00}			& 69.9 		& 49.1 			& 434 \\ \hline
		N$_7$ & \textbf{91.27} 	& 90.86						& 29.5 		& 92.2 			& 197 \\ \hline
	\end{tabular}
	\label{tab:networks}
\end{table}

Table $2$ shows the patch classification accuracy $A$ (based on the validation data $\{ P_i^{\text{val}} \}_{1 \leq i \leq N}$), average required segmentation time $T$ (in seconds) for one $3069 \times 2301$ slice of our dataset, percentage of pruned parameters $\Delta P$ and network memory usage $M$ for a batch size of $1$ (in kilobytes) for all the networks. We compare our proposed loss-based pruning method to the sparsity-based approach, proposed in \cite{Han2015}. Note we obtain competitive (and in some cases even better) performance compared to the latter method. The pruned networks N$_i$ ($i=1,\dots,6$) require $9$ to $39\%$ less segmentation time and $10$ to $34\%$ less network memory at the cost of at most $0.3\%$ accuracy. Compared to the original network, the network N$_7$ is almost $4$ times faster with an accuracy decrease of $1.9\%$, compared to $2.3\%$ for the method proposed in \cite{Han2015}. Given that our specific SEM setup requires $60$ seconds for acquiring a $3069 \times 2301$ slice and $15$ seconds for sectioning, this means we are able to perform real-time membrane segmentation at an accuracy of at most $92.87\%$. 

\begin{figure}[t]
	\centering
	\begin{minipage}{0.48\linewidth}
		\includegraphics[width=\textwidth]{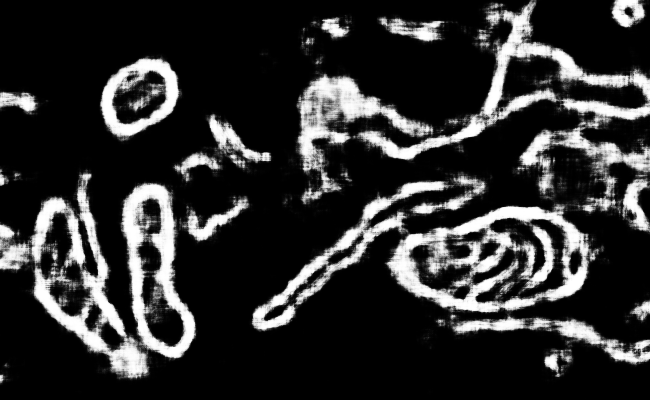}
		\centerline{(a) N}
	\end{minipage}
	\begin{minipage}{0.48\linewidth}
		\includegraphics[width=\textwidth]{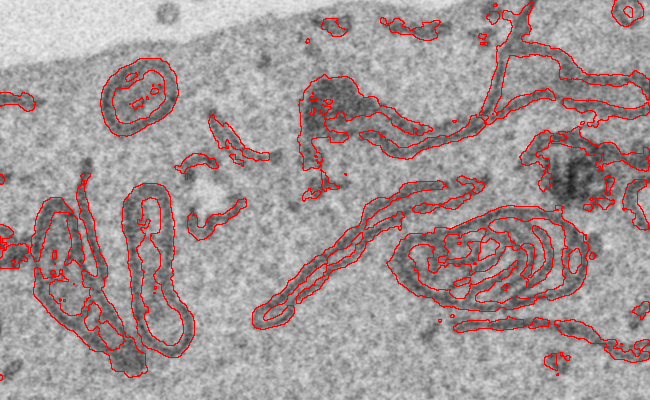}
		\centerline{(b) N} 
	\end{minipage}
	\begin{minipage}{0.48\linewidth}
		\includegraphics[width=\textwidth]{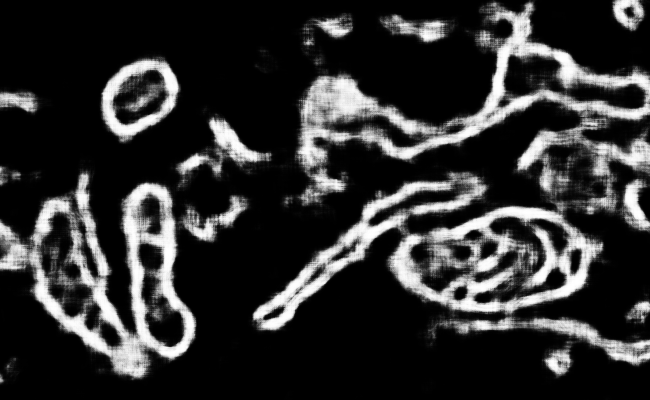}
		\centerline{(c) N7 (proposed)}
	\end{minipage}
	\begin{minipage}{0.48\linewidth}
		\includegraphics[width=\textwidth]{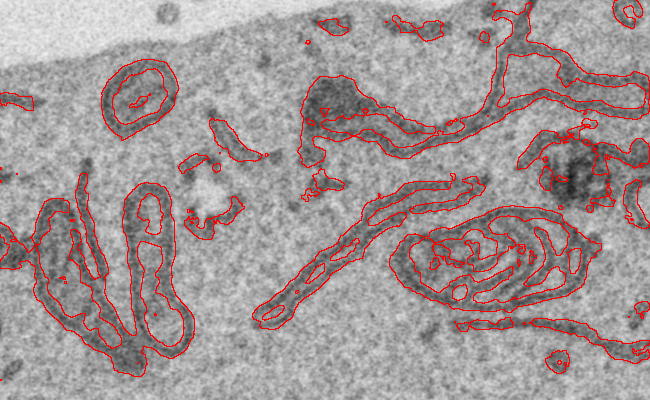}
		\centerline{(d) N7 (proposed)}
	\end{minipage}
	\begin{minipage}{0.48\linewidth}
		\includegraphics[width=\textwidth]{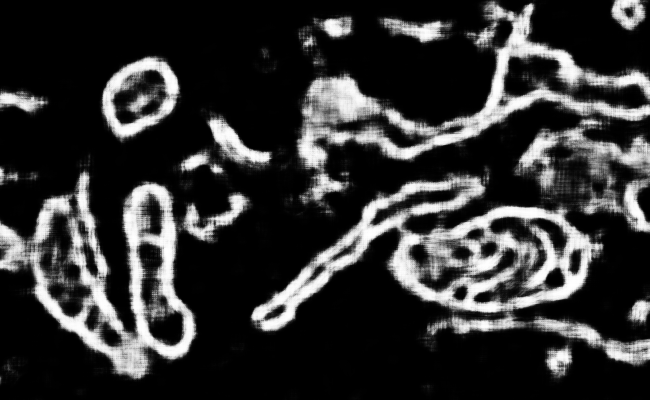}
		\centerline{(e) N7 (sparsity-based \cite{Han2015})}
	\end{minipage}
	\begin{minipage}{0.48\linewidth}
		\includegraphics[width=\textwidth]{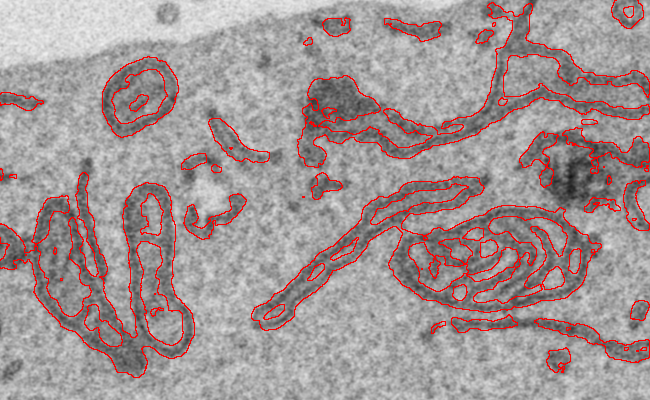}
		\centerline{(f) N7 (sparsity-based \cite{Han2015})}
	\end{minipage}
	\caption{Resulting probability maps (left) after pixel-based forward propagation of the networks N and N7 on an EM image. The result obtained by our proposed pruning method and \cite{Han2015} are both shown. The probability maps are thresholded in order to obtain the final membrane segmentation (right). }
	\label{fig:segmentation}
\end{figure}

The corresponding probability maps and segmentation result of the original network N and the stripped network N$_7$ (both our proposed loss-based pruning and the method proposed in \cite{Han2015}) are shown in Fig. \ref{fig:segmentation}. The probability maps corresponding to the N$_7$ pruned network are very similar to the one obtained by the network N. Due to the accuracy drop of approximately $2\%$, we note that the original network is able to delineate the membranes more closely, whereas the N$_7$ probability map is more blurred. When comparing the pruned network, we see that the result obtained by sparsity-based pruning is slightly more noisy than our approach.

\section{CONCLUSIONS}
\label{sec:ref}
Membrane segmentation can be a helpful tool in segmenting cell content in electron microscopy (EM) images. State-of-the-art methods such as convolutional neural networks, require large amounts of time and cannot keep up with the high throughput nature of EM. Therefore, we have proposed a convolutional neural network that achieves $93\%$ pixel classification accuracy and a pruning method that minimizes the training loss function of the network. This way, one of our pruned networks was able to accelerate the original network by a factor $4$ at a limited accuracy cost of $1.9\%$. Additionally, we have shown that our loss-based pruning method is competitive with a state-of-the-art sparsity-based method. Future work consists of network optimization, further feature analysis and reduction, instance selection and transformation in order to train neural networks more efficiently.

\bibliographystyle{IEEEbib}
\bibliography{refs}

\end{document}